\documentclass[runningheads]{llncs}

\usepackage{eccv}

\usepackage{eccvabbrv}

\usepackage{graphicx}
\usepackage{booktabs}
\usepackage{tabularx}

\usepackage[accsupp]{axessibility}  

\usepackage{hyperref}

\usepackage{orcidlink}
\usepackage{makecell}
\usepackage{array}
\usepackage{wrapfig}
\usepackage{float}

\usepackage{hhline} 
\usepackage{amssymb} 
\usepackage{pifont}

\newcommand{\xmark}{\text{\ding{55}}}

\newcolumntype{M}[1]{>{\centering\arraybackslash}m{#1}}

\usepackage{arydshln}
\definecolor{lightblue}{rgb}{0.88, 0.95, 0.97}

\newcolumntype{Y}{>{\centering\arraybackslash\hsize=.6\hsize}X}
\newcolumntype{Z}{>{\centering\arraybackslash\hsize=1\hsize}X}

\renewcommand{\thefootnote}{\fnsymbol{footnote}}

\makeatletter
\def\@fnsymbol#1{\ensuremath{\ifcase#1\or *\or \dagger\else\@ctrerr\fi}}

\newcommand{\equalcontribution}{\thanks{These authors contributed equally to this work.}}
\newcommand{\correspondingauthor}{\thanks{Corresponding author. Email: whyu@dlut.edu.cn}}

\begin{document}
\title{SGS-SLAM: Semantic Gaussian Splatting For Neural Dense SLAM} 


\author{
Mingrui Li \inst{1}\equalcontribution \and
Shuhong Liu \inst{2}$^*$ \and
Heng Zhou \inst{3} \and
Guohao Zhu \inst{2} \and
Na Cheng \inst{1} \and
Tianchen Deng \inst{4} \and
Hongyu Wang \inst{1}\correspondingauthor
}

\authorrunning{M.~Li and S.~Liu et al.}


\institute{
  Dalian University of Technology ~\and The University of Tokyo \\
  \and Columbia University ~\and Shanghai Jiao Tong University
}

\maketitle
\def\thefootnote{~}\footnotetext{The demo video is available at \url{https://youtu.be/y83yw1E-oUo}.}

\begin{abstract}
    We present SGS-SLAM, the first semantic visual SLAM system based on Gaussian Splatting. It incorporates appearance, geometry, and semantic features through multi-channel optimization, addressing the oversmoothing limitations of neural implicit SLAM systems in high-quality rendering, scene understanding, and object-level geometry. We introduce a unique semantic feature loss that effectively compensates for the shortcomings of traditional depth and color losses in object optimization. Through a semantic-guided keyframe selection strategy, we prevent erroneous reconstructions caused by cumulative errors. Extensive experiments demonstrate that SGS-SLAM delivers state-of-the-art performance in camera pose estimation, map reconstruction, precise semantic segmentation, and object-level geometric accuracy, while ensuring real-time rendering capabilities. The implementation code is available at \url{https://github.com/ShuhongLL/SGS-SLAM}.
\end{abstract}


\section{Introduction}

\begin{figure}[tb]
    \begin{center}
        \includegraphics[width=0.8\textwidth]{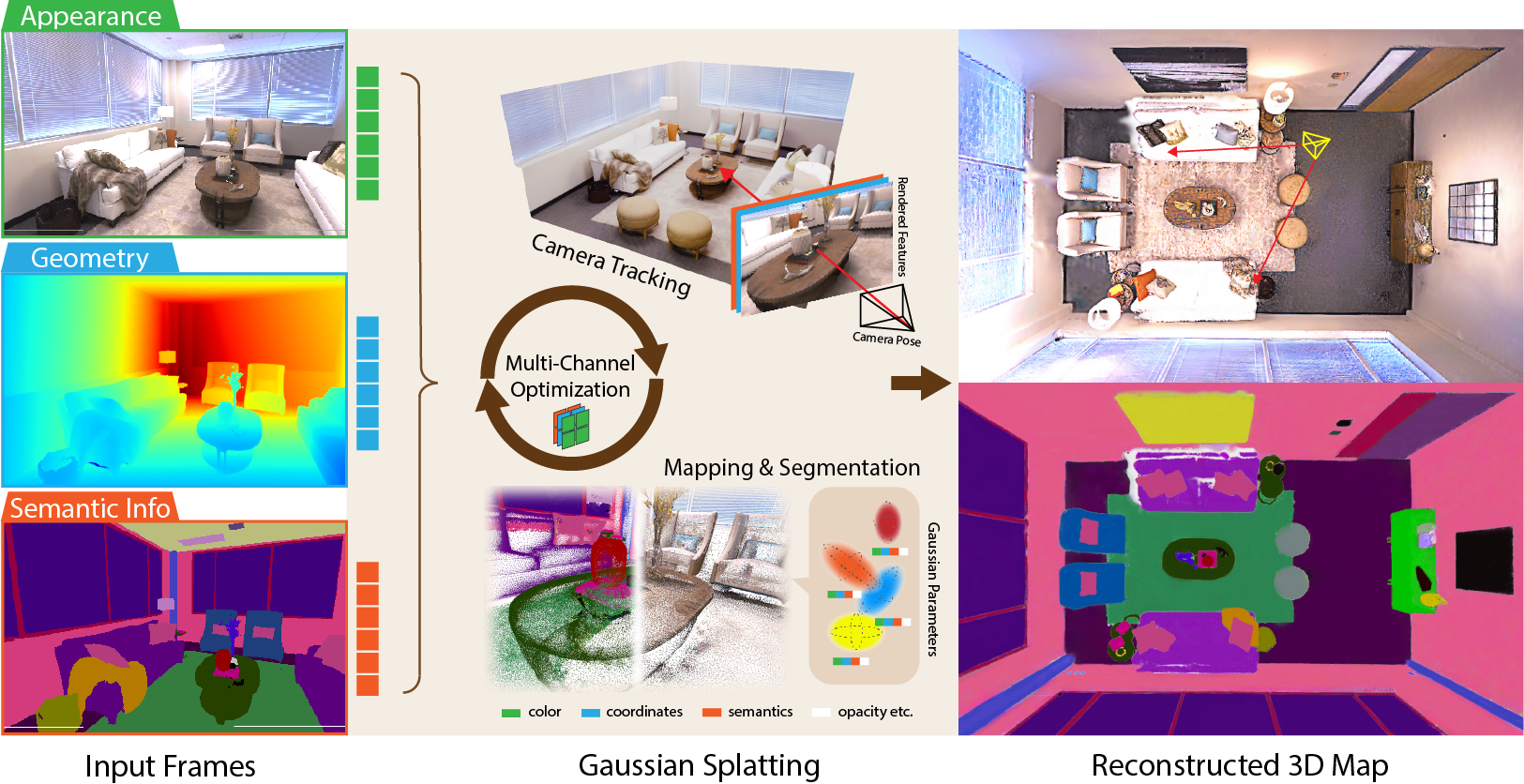}
    \end{center}
    \caption{The illustration of the proposed SGS-SLAM. It employs 2D inputs encompassing appearance, geometry, and semantic information, leveraging Gaussian Splatting and differentiable rendering for multi-channel parameter optimization. During the mapping process, SGS-SLAM maps the 2D semantic prior to the 3D scene, jointly optimizing it via the mapping loss for accurate 3D segmentation outcomes.}
    \label{fig:abstract}
\end{figure}

Dense Visual Simultaneous Localization and Mapping (SLAM) is a crucial problem in the field of computer vision. It aims to reconstruct a dense 3D map in an unseen environment while simultaneously tracking the camera poses. Traditional visual SLAM systems \cite{davison2007monoslam, newcombe2011dtam, salas2013slam, mur2015orb} stand out in sparse mapping using point clouds and voxels, but fall short in dense reconstruction. To extract dense geometric information for high-quality representation, learning-based SLAM methods \cite{bloesch2018codeslam, sucar2020nodeslam} have gained wild attention. They demonstrate proficiency in generating decent 3D global maps meanwhile exhibiting robustness on noises and outliers. Drawing inspiration from advancements in the neural radiance field (NeRF) \cite{mildenhall2021nerf}, NeRF-based SLAM approaches \cite{sucar2021imap, zhu2022nice, kong2023vmap, zhang2023go, li2023dns, wang2023co, deng2024plgslam} have made further progress. They excel in producing accurate and high-fidelity global reconstruction by capturing dense photometric information through differentiable rendering.

However, NeRF-based SLAM methods employ multi-layer perceptrons (MLPs) as the implicit neural representation of scenes, which introduces several challenging limitations. Primarily, MLP models struggle with over-smoothing issues at the edge of objects, leading to a lack of fine-grained details in the map. This challenge also brings difficulties in disentangling the representation of objects, making it non-trivial to segment, edit, and manipulate objects within the scene. Moreover, when applied to larger scenes, MLP models are prone to catastrophic forgetting. This means that incorporating new scenes can adversely affect the precision of previously learned models, thereby reducing overall performance. Additionally, NeRF-based methods are computationally inefficient. Since the entire scene is modeled through one or several MLPs, it necessitates extensive model tuning for adding or updating scenes.


In this context, as opposite to NeRF-based neural representation, our exploration shifts towards the volumetric representation based on the 3D Gaussian Radiance Field \cite{kerbl20233d}. This approach marks a significant shift and offers notable advantages in the scene representation.

Benefits from its rasterization of 3D primitives, Gaussian Splatting exhibits remarkably fast rendering speeds and allows direct gradient flow to each Gaussian's parameters. This results in an almost linear projection between the dense photometric loss and parameters during optimization \cite{keetha2023splatam}, unlike the hierarchical pixel sampling and indirect gradient flow through multiple non-linear layers seen in NeRF models. Moreover, the direct projection capability simplifies the addition of new channels to the Gaussian field, thereby enabling dynamic multi-channel feature rendering. Crucially, we integrate a semantic map into the 3D Gaussian field, essential for applications in robotics and mixed reality. This integration allows real-time rendering of appearance, depth, and semantic color. 


When compared with neural implicit semantic SLAM systems, such as DNS-SLAM \cite{li2023dns} and SNI-SLAM \cite{zhu2023sni}, our system demonstrates remarkable superiority in terms of rendering speed, reconstruction quality, and segmentation accuracy. Leveraging these benefits, our method enables precise editing and manipulation of specific scene elements while preserving the high fidelity of the overall rendering. Furthermore, using explicit spatial and semantic information to identify scene content can be instrumental in optimizing camera tracking. Particularly, we incorporate a two-level adjustment based on geometric and semantic criteria for keyframes selection. This process relies on recognizing objects that have been previously seen in the trajectory. Extensive experiments are conducted on both synthetic and real-world scene benchmarks. These experiments compare our method against both implicit NeRF-based approaches \cite{yang2022vox,zhu2022nice,wang2023co,johari2023eslam}, and novel 3D-Gaussian-based methods \cite{keetha2023splatam}, evaluating performance in mapping, tracking, and semantic segmentation.

Overall, our work presents several key contributions, summarized as follows:
\begin{itemize}
\item We introduce SGS-SLAM, the first semantic RGB-D SLAM system grounded in 3D Gaussians. SGS-SLAM employs an explicit volumetric representation, enabling swift and real-time camera tracking and scene mapping. More importantly, it utilizes 2D semantic maps to learn 3D semantic representations expressed by Gaussians. Compared with previous NeRF-based methods which offer over-smooth object edges, SGS-SLAM provides high-fidelity reconstruction and optimal segmentation precision. 

\item In SGS-SLAM, semantic maps provide additional supervision for optimizing parameters and selecting keyframes. We employ a multi-channel parameter optimization strategy where appearance, geometric, and semantic signals collectively contribute to camera tracking and scene reconstruction. Furthermore, SGS-SLAM utilizes these diverse channels for keyframe selection during the tracking phase, concentrating on actively recognizing objects seen earlier in the trajectory.


\item Utilizing the semantic map, SGS-SLAM provides a highly accurate disentangled object representation in 3D scenes, laying a solid foundation for downstream tasks such as scene editing and manipulation. SGS-SLAM facilitates the dynamic moving, rotating, or removal of objects that is achieved by grouping Gaussians by specifying the semantic labels of the objects.

\end{itemize}

\section{Related Work}

\subsection{Semantic SLAM}
Semantic information is of great importance for SLAM systems \cite{mur2015orb, he2023ovd, whelan2015elasticfusion, qin2018vins}, which is a crucial requirement for applications in robotics and VR or AR fields. Real-time dense semantic SLAM systems \cite{salas2013slam, bloesch2018codeslam, rosinol2020kimera} integrate semantic information into 3D geometric representations. Traditional semantic SLAM systems rely on sparse 3D semantic expressions, such as voxel \cite{hermans2014dense}, point cloud \cite{narita2019panopticfusion}, and signed distance field \cite{narita2019panopticfusion}. These methods struggle with accurately interpreting complex environments due to limited semantic understanding. This results in a simplified categorization of environmental features, which may not capture the full range of objects and their relationships within a space. Moreover, these methods exhibit limitations regarding reconstruction speed, high-fidelity model acquisition, and memory usage.

\subsection{Neural Implicit SLAM}
Methods based on NeRF \cite{mccormac2018fusion}, which handle complex topological structures and differentiable scene representation methods, have garnered significant attention, leading to the development of neural implicit SLAM methods \cite{chung2023orbeez, deng2023long, li2023end, li2024ddn, deng2024incremental, zhou2024mod, deng2024neslam}. iMAP \cite{sucar2021imap} uses a single MLP for scene representation, which shows limitations in large-scale scenes. NICE-SLAM \cite{zhu2022nice} uses pre-trained multiple MLPs for hierarchical scene representation. Co-SLAM \cite{wang2023co} combines pixel-set-based keyframe tracking with one-blob encoding. Go-SLAM \cite{zhang2023go} uses Droid-SLAM \cite{teed2021droid} as the tracking system and multi-resolution hash encoding \cite{muller2022instant} for mapping. However, these methods cannot utilize semantic information in the map. NIDS-SLAM \cite{haghighi2023neural} leverages the tracking system of ORB-SLAM3 \cite{campos2021orb} and Instant-NGP \cite{muller2022instant} for mapping but does not optimize joint semantic features for 3D reconstruction. DNS-SLAM \cite{li2023dns} proposes a 2D semantic prior system that provides multi-view geometry constraints but does not optimize 3D reconstruction with semantic features. DNS-SLAM \cite{li2023dns} and SNI-SLAM \cite{zhu2023sni} introduce semantic loss for geometric supervision but remain limited by the efficiency constraints of NeRF's volume rendering.

\subsection{3D Gaussian Splatting SLAM}
Utilizing the outstanding performance and fast rasterization capabilities of 3D Gaussian Splatting \cite{kerbl20233d}, Gaussian-based SLAM systems offer higher efficiency and accuracy on scene reconstruction \cite{yan2023gs, keetha2023splatam, huang2023photo, matsuki2024gaussian, yugay2023gaussian, deng2024compact, liu2024structure}. However, existing Gaussian-based SLAM systems lack the ability to recognize semantic information in scenes. To bridge this gap, We utilize the semantic map during keyframe selection and integrate the semantic feature loss in the tracking and mapping process. This allows us to obtain more effective and higher-quality scene segmentation outcomes meanwhile preserving real-time processing performance.

\section{Method}
SGS-SLAM is a Gaussian-based semantic visual SLAM system. \cref{sec:gaussian_representation} introduces its multi-channel Gaussian representation for joint parameter optimization. Like previous SLAM techniques, our method can be split into two processes: tracking and mapping. The tracking process estimates the camera pose of each frame while keeping the scene parameters fixed. Mapping optimizes the scene representations based on the estimated camera pose. \cref{sec:tracking_mapping} explains the breakdown steps in detail. In addition, \cref{sec:scene_manipulation} presents scene manipulation as a case study for downstream tasks. \cref{fig:abstract} shows an overview of our system.

\subsection{Multi-Channel Gaussian Representation}
\label{sec:gaussian_representation}

The scene is represented using a Gaussian influence function $f(\cdot)$ on the map, For simplicity, these Gaussians are isotropic, as proposed in \cite{keetha2023splatam}:

\begin{equation}
f^{\rm 3D}(x) = \sigma \exp\left(-\frac{\|x - \mu\|^2}{2r^2}\right) \label{eq:gaussian-definition}
\end{equation}

\noindent Here, $\sigma \in [0, 1]$ indicates opacity, $\mu \in \mathbb{R}^3$ represents the center position, and $r$ denotes the radius. Each Gaussian also carries RGB colors $c_i = [r_i~b_i~g_i]^T$.

In order to optimize the parameters of Gaussians to represent the scene, we need to render the Gaussians into 2D images in a differentiable manner. We use the render method from \cite{luiten2023dynamic}, providing extended functionality of rendering depth in colors. It works by splatting 3D Gaussians into the image plane via approximating the integral projection of the influence function $f(\cdot)$ along the depth dimension in pixel coordinates. The center of the Gaussian $\mu$, radius $r$, and depth $d$ (in camera coordinates) is splatted using the standard point rendering formula:

\begin{equation}
\mu^{\rm 2D} = K \frac{E_t \mu}{d}, \quad r^{\rm 2D} = \frac{lr}{d}, \quad d = (E_t \mu)_z
\end{equation}

\noindent where $K$ is the camera intrinsic matrix, $E_t$ is the extrinsic matrix capturing the rotation and translation of the camera at frame $t$, $l$ is the focal length. The influence of all Gaussians on this pixel can be combined by sorting the Gaussians in depth order and performing front-to-back volume rendering:

\begin{equation}
C_{\rm pix} = \sum_{i=1}^{n} c_i f_{i,\rm pix}^{\rm 2D} \prod_{j=1}^{i-1} (1 - f_{j,\rm pix}^{\rm 2D})
\end{equation}

The pixel-level rendered color $C_{\rm pix}$ is the sum over the colors of each Gaussian $c_i$ and weighted by the influence function $f_{i,\rm pix}^{\rm 2D}$ (replace the 3D means and covariance matrices with the 2D splatted versions), multiplied by an occlusion term taking into account the effect of all Gaussians in front of the current Gaussian. Similarly, the depth can be rendered as:
\begin{equation}
D_{\rm pix} = \sum_{i=1}^{n} d_i f_{i,\rm pix}^{\rm 2D} \prod_{j=1}^{i-1} (1 - f_{j,\rm pix}^{\rm 2D})
\end{equation}

\noindent where $d_i$ denotes the depth of each Gaussian. By setting $d_i=1$, we can calculate a silhouette, $Sil_{\rm pix} = D_{\rm pix}(d_i=1)$, which assists in determining whether a pixel is visible in the current view \cite{keetha2023splatam}. This aspect of visibility is essential for camera pose estimation, as it relies on the current reconstructed map. Additionally, it is also employed in map reconstruction, where new Gaussians are introduced in pixels lacking sufficient information.

While acquiring 3D semantic information is challenging and usually demands extensive manual labeling, the 2D semantic label is more accessible prior. In our approach, we leverage 2D semantic labels, which are often provided in datasets or can be easily obtained using off-the-shelf methods. We assign distinct channels to the parameters of Gaussians to denote their semantic labels and colors. During the rendering process, the 2D semantic map can be rendered from the reconstructed 3D scene as follows:

\begin{equation}
S_{\rm pix} = \sum_{i=1}^{n} s_i f_{i,\rm pix}^{\rm 2D} \prod_{j=1}^{i-1} (1 - f_{j,\rm pix}^{\rm 2D})
\end{equation}

\noindent where $s_i = [r_i~b_i~g_i]^T$ denotes the semantic color associated with the Gaussian. This semantic color is optimized jointly with the appearance color and depth during the mapping process.

The Gaussian representations employed in SGS-SLAM facilitate high-quality reconstructions at high rendering speed, offering exceptional accuracy in capturing complex textures and geometry with remarkable detail and efficiency. Furthermore, the integration of semantic features within our method significantly advances optimal scene interpretation and precise object-level geometry, effectively mitigating the oversmoothing issues prevalent in NeRF models.

\subsection{Tracking and Mapping}
\label{sec:tracking_mapping}

\subsubsection{Camera Pose Estimation}

Given the first frame, the camera pose is set to identity and used as the reference coordinates for the following tracking and mapping procedure. While assessing the camera pose of an RGB-D view at a new timestep, the initial camera pose is determined by adding a displacement to the previous pose, assuming constant velocity, as $E_{t+1} = E_t + (E_{t} - E_{t-1})$. Following this, the current pose is iteratively refined by minimizing the tracking loss between the ground truth color ($C_{\rm pix}^{GT}$), depth images ($D_{\rm pix}^{GT}$), and semantic map ($S_{\rm pix}^{GT}$) and their differentiably rendered views:

\begin{equation}
\mathcal{L}_{\rm tracking} = \sum_{\rm pix} (Sil_{\rm pix} > T_{\rm sil})(\lambda_D |D_{\rm pix}^{GT} - D_{\rm pix}| + \lambda_C |C_{\rm pix}^{GT} - C_{\rm pix}| + \lambda_S |S_{\rm pix}^{GT} - S_{\rm pix}|)
\end{equation}

Here, only those rendered pixels with a sufficiently large silhouette are factored into the loss calculation. The threshold $T_{\rm sil}$ is designed to make use of the map that has been previously optimized and has high certainty to be visible in the current camera view.

\subsubsection{Keyframes Selection and Weighting}
\label{section:key-frame-selection}
During the tracking phase of SLAM systems, keyframes are identified and stored simultaneously. These keyframes, providing different views of objects, are critical for mapping to refine 3D scene reconstruction. SGS-SLAM captures and stores keyframes at constant time intervals. Subsequently, keyframes associated with the current frame are chosen based on geometric and semantic constraints. Specifically, we randomly select pixels from the current frame and extract their corresponding Gaussians $G_{\rm sample}$ in the 3D scene. These sampled $G_{\rm sample}$ are then projected onto the camera views of keyframes as $G_{\rm proj}$, which are evaluated based on the geometric overlap ratio:

\begin{equation}
\eta = \frac{1}{\sum_{} G_{\rm proj}} \sum_{n=i} \{G_i | 0 \leq width(G_i) \leq W, 0 \leq height(G_i) \leq H\}
\end{equation}

It represents the proportion of Gaussians captured within the camera view of the keyframes. $W$ and $H$ are the width and height of the camera view. The candidates with $\eta$ lower than a certain threshold $T_{\rm geo}$ are removed. After the initial geometric-based selection, a second filter is conducted based on semantic criteria. We discard keyframes whose semantic maps $S_{\rm pix}$ are identical to the current frame's semantic map, as indicated by a high mIoU score. This threshold, denoted as $T_{\rm sem}$ intends to enhance map optimization from varying viewpoints, preferring views with low mIoU overlap. The remaining candidates are randomly sampled to serve as the selected keyframes associated with the current frame. In addition, we compute an uncertainty score for each keyframe, defined as $\mathcal{U}(t) = e^{-\tau t}$, with $t$ representing the timestamp of the keyframe and $\tau$ being a decay coefficient. This uncertainty score is used to weight the mapping loss $\mathcal{L}_{\rm mapping}$. The intuition behind this is that keyframes with a later timestamp index carry a higher uncertainty in reconstruction due to the accumulation of camera tracking errors along the trajectory.

\subsubsection{Map Reconstruction}

The scene is modeled using Gaussians across three distinct channels: (1) their mean coordinates represent the geometric information of the scene, (2) their appearance colors depict the scene's visual appearance, and (3) their semantic colors indicate the semantic labels of objects. These parameters across the channels are jointly optimized during the process of Gaussian densification and optimization, whereas the camera pose, ascertained from tracking, remains fixed.

Starting with the first frame, all pixels contribute to initializing the map. In the process of map reconstruction at a new timestep, new Gaussians are introduced to areas of the map that are either insufficiently dense or display new geometry in front of the previously estimated map. The addition of new Gaussians is regulated by applying a mask to the pixels where either (ii) the silhouette value $Sil_{\rm pix}$ falls below a certain threshold, signifying a high uncertainty in visibility, or (ii) the ground-truth depth is much smaller than the estimated depth, suggesting the presence of new geometric entities.

After densification, the parameters of the map are optimized by minimizing the mapping loss:

\begin{equation}
\mathcal{L}_{\rm mapping} = \mathcal{U}_t \sum_{\rm pix} \lambda_D |D_{\rm pix}^{GT} - D_{\rm pix}| + \lambda_C \mathcal{L}_C + \lambda_S \mathcal{L}_S
\end{equation}

\noindent where $\mathcal{L}_C$ and $\mathcal{L}_S$ are weighted SSIM loss \cite{kerbl20233d} with respect to appearance image and semantic image:

\begin{equation}
\mathcal{L}(I_{\rm pix}) = \sum_{\rm pix} \alpha |I_{\rm pix}^{GT} - I_{\rm pix}| + (1-\alpha)(1 - ssim(I_{\rm pix}^{GT}, I_{\rm pix}))
\end{equation}

\noindent $\lambda_D$, $\lambda_C$, $\lambda_S$, and $\alpha$ are predefined hyperparameters, and $\mathcal{U}_t$ is the uncertainty score defined in \cref{section:key-frame-selection}.

Compared to existing NeRF-based approaches \cite{zhu2022nice,johari2023eslam,li2023dns,zhu2023sni} that necessitate complex model architectures and feature fusion strategies, SGS-SLAM adopts explicit Gaussian representation for mapping, resulting in high rendering speeds and optimal reconstruction quality. Compared to recent Gaussian-based methods \cite{keetha2023splatam, yan2023gs}, SGS-SLAM incorporates geometric, appearance, and semantic features for multi-channel rendering. This enables the joint optimization of parameters across different channels, remarkably enhancing the efficiency and effectiveness of both mapping and segmentation processes.


\subsection{Scene Manipulation via Object-level Geometry}
\label{sec:scene_manipulation}
Given that the scene is represented explicitly by Gaussians, it becomes feasible to directly edit and manipulate a targeted group of Gaussians. In our case, Gaussian groups are identified based on their semantic labels. The mapping process generates these Gaussians, as defined in \cref{eq:gaussian-definition}, allowing for further manipulation in the following manner:

\begin{equation}
f^{\rm 3D}_{\rm edit}(G, \tilde{y}) = M(G, \tilde{y}) \cdot \Phi_T(f^{\rm 3D}(G), \tilde{y}) \label{eq:scene_manipulation}
\end{equation}

\noindent where the edited Gaussians, $f^{\rm 3D}_{\rm edit}$, are influenced by the visibility mask $M$, transition function $\Phi_T$, and the Gaussian's semantic label $\tilde{y}$. The visibility mask $M$ determines if the Gaussians should be retained (1) or removed (0) based on $\tilde{y}$. The transition function $\Phi_T$ applies a transformation to the Gaussian's coordinates on selected $\tilde{y}$, enabling spatial manipulation.

\section{Experiment}

\subsection{Experimental Setup}

\subsubsection{Datasets}
We evaluate our method on both synthetic and real-world datasets. To compare with other neural implicit SLAM methods, we evaluate synthetic scenes from Replica dataset \cite{straub2019replica} and real-world scenes from ScanNet \cite{dai2017scannet} and ScanNet++ \cite{yeshwanthliu2023scannetpp} datasets. The ground-truth camera pose and semantic map of Replica are offered from simulation, and the ground-truth camera pose of ScanNet is generated by BundleFusion \cite{dai2017bundlefusion}. The ground-truth 2D semantic label is provided by the dataset.

\subsubsection{Metrics}
We use PSNR, Depth-L1, SSIM, and LPIPS to evaluate the reconstruction quality. To evaluate the camera pose, we use the average absolute trajectory error (ATE RMSE). For semantic segmentation, we calculate mIoU score.

\subsubsection{Baselines}
We compare the tracking and mapping with state-of-the-art methods NICE-SLAM \cite{zhu2022nice}, vMap \cite{kong2023vmap}, Co-SLAM \cite{wang2023co}, ESLAM \cite{johari2023eslam}, and SplaTAM \cite{keetha2023splatam}. For semantic segmentation accuracy, we compare with NIDS-SLAM \cite{haghighi2023neural}, DNS-SLAM \cite{li2023dns}, and SNI-SLAM \cite{zhu2023sni}.

\begin{figure}[tb]
    \begin{center}
        \includegraphics[width=0.85\textwidth]{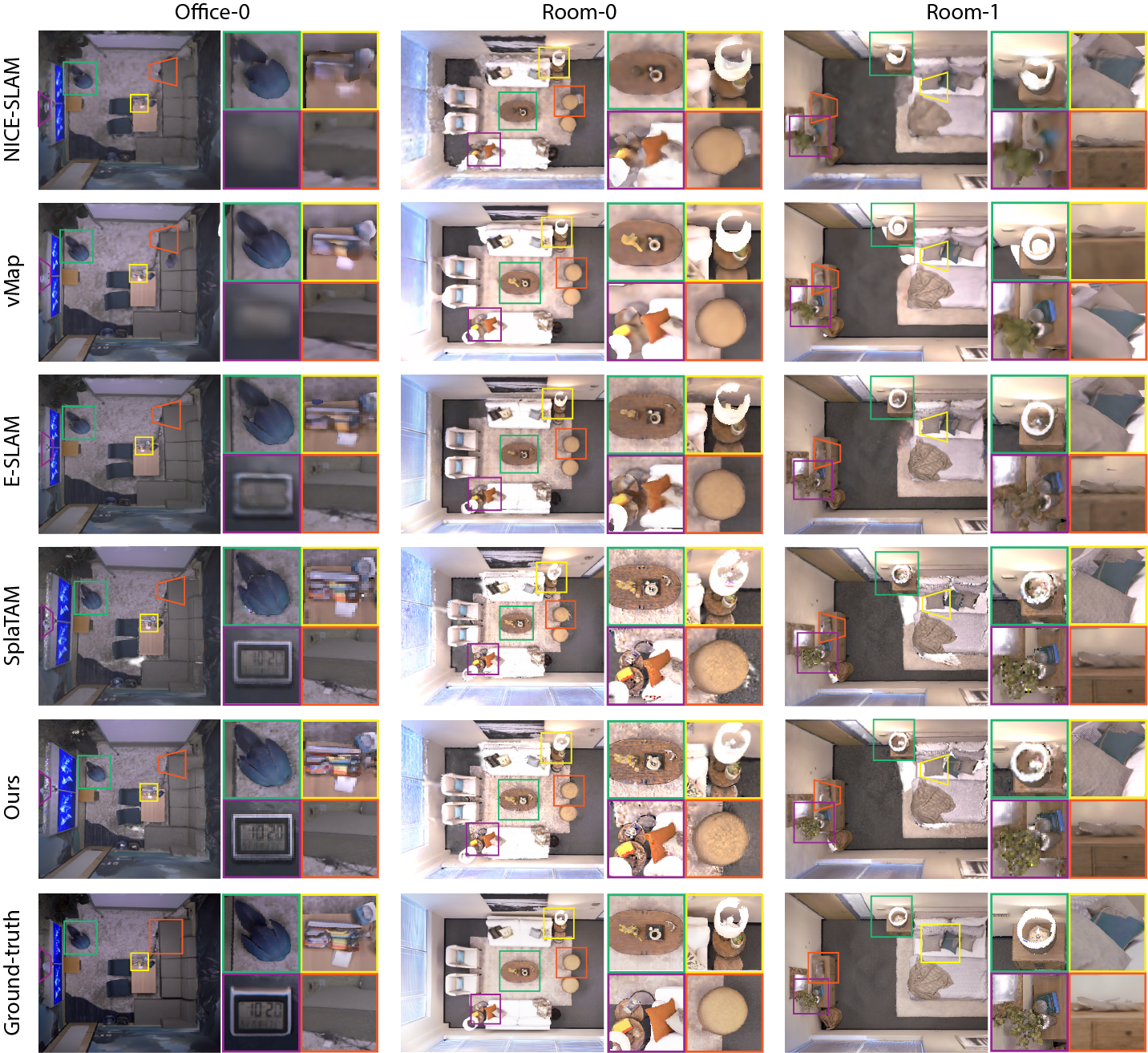}
    \end{center}
    \caption{Qualitative comparison of our method and the baselines for reconstruction across three scenes from the Replica Dataset \cite{straub2019replica}, with key details accentuated using colored boxes. The results demonstrate that our method delivers more high-fidelity and robust reconstructions.}
    \label{fig:demo}
\end{figure}

\subsection{Evaluation of Mapping and Localization}
We show quantitative measures of reconstruction quality using the Replica dataset \cite{straub2019replica} in \cref{tab:psnr}. Our method demonstrates state-of-the-art performance. Compared to other baseline methods, our approach attains notably superior outcomes, outperforming them by a margin of 10dB in PSNR.

\begin{table}[tb]
    \centering
    \caption{Quantitative comparison of our method and the baselines in training view rendering on the Replica dataset \cite{straub2019replica}. Our method demonstrates \colorbox{lightblue}{SOTA} performances in most cases among three metrics.}
    \scriptsize 
    \begin{tabularx}{\textwidth}{@{}l@{\hspace{1pt}} | *{11}{>{\centering\arraybackslash}X}@{}}
        \Xhline{2\arrayrulewidth}
        \textbf{Methods} & \textbf{Metrics} & \textbf{Avg.} & Room0 & Room1 & Room2 & Office0 & Office1 & Office2 & Office3 & Office4 \\    
        \hline
         & PSNR$\uparrow$ & 24.42 & 22.12 & 22.47 & 24.52 & 29.07 & 30.34 & 19.66 & 22.23 & 24.94 \\
        NICE-SLAM & SSIM$\uparrow$ & 0.809 & 0.689 & 0.757 & 0.814 & 0.874 & 0.886 & 0.797 & 0.801 & 0.856 \\
         & LPIPS$\downarrow$ & 0.233 & 0.330 & 0.271 & 0.208 & 0.229 & 0.181 & 0.235 & 0.209 & 0.198 \\
        \hline
        & PSNR$\uparrow$ & 30.24 & 27.27 & 28.45 & 29.06 & 34.14 & 34.87 & 28.43 & 28.76 & 30.91 \\
        Co-SLAM & SSIM$\uparrow$ & 0.939 & 0.910 & 0.909 & 0.932 & 0.961 & 0.969 & 0.938 & 0.941 & 0.955 \\
        & LPIPS$\downarrow$ & 0.252 & 0.324 & 0.294 & 0.266 & 0.209 & 0.196 & 0.258 & 0.229 & 0.236 \\
        \hline
        & PSNR$\uparrow$ & 29.08 & 25.32 & 27.77 & 29.08 & 33.71 & 30.20 & 28.09 & 28.77 & 29.71 \\
        ESLAM & SSIM$\uparrow$ & 0.929 & 0.875 & 0.902 & 0.932 & 0.960 & 0.923 & 0.943 & 0.948 & 0.945 \\
        & LPIPS$\downarrow$ & 0.336 & 0.313 & 0.298 & 0.248 & 0.184 & 0.228 & 0.241 & 0.196 & 0.204 \\
        \hline
        & PSNR$\uparrow$ & 33.98 & 32.48 & 33.72 & 34.96 & 38.34 & 39.04 & 31.90 & 29.70 & 31.68 \\
        SplaTAM & SSIM$\uparrow$ & 0.969 & 0.975 & 0.970 & \cellcolor{lightblue}\textbf{0.982} & 0.982 & \cellcolor{lightblue}\textbf{0.982} & 0.965 & 0.950 & 0.946 \\
        & LPIPS$\downarrow$ & 0.099 & 0.072 & 0.096 & 0.074 & \cellcolor{lightblue}\textbf{0.083} & 0.093 & \cellcolor{lightblue}\textbf{0.100} & 0.118 & 0.155 \\
        \hline
        \noalign{\vskip 0.2pt}
        & PSNR$\uparrow$ & \cellcolor{lightblue}\textbf{34.66} & \cellcolor{lightblue}\textbf{32.50} &
        \cellcolor{lightblue}\textbf{34.25} &
        \cellcolor{lightblue}\textbf{35.10} &
        \cellcolor{lightblue}\textbf{38.54} &
        \cellcolor{lightblue}\textbf{39.20} &
        \cellcolor{lightblue}\textbf{32.90} &
        \cellcolor{lightblue}\textbf{32.05} &
        \cellcolor{lightblue}\textbf{32.75} \\
        \textbf{Ours} & SSIM$\uparrow$ &
        \cellcolor{lightblue}\textbf{0.973} &
        \cellcolor{lightblue}\textbf{0.976} &
        \cellcolor{lightblue}\textbf{0.978} &
        0.981 &
        \cellcolor{lightblue}\textbf{0.984} &
        0.980 &
        \cellcolor{lightblue}\textbf{0.967} &
        \cellcolor{lightblue}\textbf{0.966} &
        \cellcolor{lightblue}\textbf{0.949} \\
        & LPIPS$\downarrow$ &
        \cellcolor{lightblue}\textbf{0.096} &
        \cellcolor{lightblue}\textbf{0.070} &
        \cellcolor{lightblue}\textbf{0.094} &
        \cellcolor{lightblue}\textbf{0.070} &
        0.086 &
        \cellcolor{lightblue}\textbf{0.087} &
        0.101 &
        \cellcolor{lightblue}\textbf{0.115} &
        \cellcolor{lightblue}\textbf{0.148} \\
        \Xhline{2\arrayrulewidth}
    \end{tabularx}
    \label{tab:psnr}
\end{table}

\begin{table}[tb]
    \centering
    \caption{Quantitative comparison in terms of Depth L1, ATE, and FPS between our method and the baselines on the Replica dataset \cite{straub2019replica}. The values represent the average outcomes across eight scenes. The FPS is evaluated by setting the same number of training iterations for all systems for fair comparison. The results of baselines are retrieved from \cite{zhu2023sni}. Our method outperforms the baselines at Depth and ATE evaluations, and performs fairly on FPS metrics. \colorbox{lightblue}{SOTA} performances are highlighted.}
    \scriptsize 
    \begin{tabularx}{\textwidth}{X | *{6}{>{\centering\arraybackslash}X}}
    \Xhline{2\arrayrulewidth}
    \textbf{Methods} & \makecell{Depth L1 \\ {[cm]$\downarrow$}} & \makecell{ATE Mean \\ {[cm]$\downarrow$}} & \makecell{ATE RMSE \\ {[cm]$\downarrow$}} &  \makecell{Track. FPS \\ {[f/s]$\uparrow$}} & \makecell{Map. FPS \\ {[f/s]$\uparrow$}} & \makecell{SLAM FPS \\ {[f/s]$\uparrow$}} \\ 
    \hline
    NICE-SLAM & 1.903 & 1.795 & 2.503 & 13.70 & 0.20 & 0.20 \\
    Co-SLAM & 1.513 & 0.935 & 1.059 & 17.24 & \cellcolor{lightblue}\textbf{10.20} & \cellcolor{lightblue}\textbf{6.41} \\
    ESLAM & 1.180 & 0.520 & 0.630 & \cellcolor{lightblue}\textbf{18.11} & 3.62 & 3.02 \\
    SplaTAM & 0.525 & 0.348 & 0.454 & 5.53 & 3.84 & 2.26 \\
    \textbf{Ours} & \cellcolor{lightblue}\textbf{0.356} & \cellcolor{lightblue}\textbf{0.327} & \cellcolor{lightblue}\textbf{0.412} & 5.27 & 3.52 & 2.11 \\
    \Xhline{2\arrayrulewidth}
    \end{tabularx}
    \label{tab:depth_l1}
\end{table}

In \cref{fig:demo}, we present the reconstruction results of three chosen scenes, where regions of interest are accentuated with boxes in various colors. Our method exhibits high-fidelity reconstruction outcomes. Specifically, for small, intricately textured objects like a clock, socket, books on a tea table, and a lamp, our approach shows remarkable accuracy over NeRF-based methods. This is because Gaussians are capable of representing objects with complex textures and surfaces. Furthermore, NeRF-based methods often struggle with the over-smoothing issue, resulting in blurred edges on objects. In contrast, by utilizing an explicit Gaussian representation, SGS-SLAM precisely captures objects with clear edges, irrespective of their sizes. Compared with SplaTAM \cite{keetha2023splatam}, which is also a Gaussian-based model, our approach utilizes semantic information for discerning object categories, recognizing visual appearance to determine texture, and applying geometric constraints to preserve accurate shapes. This combination enables our method to achieve thorough modeling of both objects and their surrounding environment. The combination of these constraints allows SGS-SLAM to capture fine-grained details of objects, offering high-fidelity and accurate reconstruction.

\cref{tab:depth_l1} displays the tracking evaluation results on the Replica dataset \cite{straub2019replica}. Our method excels in achieving the highest level of depth L1 loss (cm) and minimal ATE error, surpassing baseline methods by 70\% in terms of depth loss and 34\% in terms of ATE RMSE (cm). This exceptional performance can be attributed to our precise scene reconstruction, which provides finely-detailed rendering results. The high-quality rendering, in turn, contributes to accurate camera pose estimation based on the established map by preventing incorrect geometric reconstruction, which could otherwise result in inaccurate tracking outcomes. Additionally, utilizing features from different channels of Gaussians, such as geometry, appearance, and semantic information, provides multiple levels of supervision, resulting in a more robust and accurate tracking capability.

\subsection{Evaluation of Semantic Segmentation}

\cref{tab:miou} shows a quantitative evaluation of our method in comparison to other neural semantic SLAM approaches. It's worth noting that we only show four scenes because previous NeRF-based semantic models only reported results on these scenes. In comparison to these previous methods, SGS-SLAM demonstrates state-of-the-art performance, outperforming the initial baseline by more than 10\%. Substantial enhancement highlights the crucial advantage of explicit Gaussian representation over NeRF-based approaches. Gaussians can precisely isolate object boundaries, resulting in highly accurate 3D scene segmentation. In contrast, NeRF-based methods often struggle to recognize individual objects and typically require complex muti-level model designs and extensive feature fusion. Our approach offers an unparalleled ability to identify 3D objects in decomposed representations, which can serve as 3D priors for tracking and mapping in future time steps, and is well-suited for further downstream tasks.

\begin{table}[h]
    \centering
    \caption{Quantitative comparison of our method against existing semantic NeRF-based SLAM methods on the Replica dataset \cite{straub2019replica}. The baselines are limited to four scenes as their results are reported only for these. For each scene, we compute the average mIoU score by comparing the rendered and the ground-truth 2D semantic image in the training view. Our method significantly outperforms the NeRF-based approaches, achieving \colorbox{lightblue}{SOTA} mIoU scores over 90\%. }
    \scriptsize 
    \begin{tabularx}{\textwidth}{@{}l@{\hspace{4pt}} | @{\hspace{1pt}}*{5}{>{\centering\arraybackslash}X}@{}}
        \Xhline{2\arrayrulewidth}
        \textbf{Methods} & \textbf{Avg. mIoU}$\uparrow$ & Room0 [\%]$\uparrow$ & Room1 [\%]$\uparrow$ & Room2 [\%]$\uparrow$ & Office0 [\%]$\uparrow$ \\ 
        \hline
        NIDS-SLAM & 82.37 & 82.45 & 84.08 & 76.99 & 85.94 \\
        DNS-SLAM & 84.77 & 88.32 & 84.90 &  81.20 & 84.66 \\
        SNI-SLAM & 87.41 & 88.42 & 87.43 & 86.16 & 87.63 \\
        \textbf{Ours} & \cellcolor{lightblue}\textbf{92.72} & \cellcolor{lightblue}\textbf{92.95} & \cellcolor{lightblue}\textbf{92.91} & \cellcolor{lightblue}\textbf{92.10} & \cellcolor{lightblue}\textbf{92.90} \\
        \Xhline{2\arrayrulewidth}
    \end{tabularx}
    \label{tab:miou}
\end{table}

\subsection{Evaluation of Keyframe Optimization}

\begin{figure}[!t]
    \begin{center}
        \includegraphics[width=0.8\textwidth]{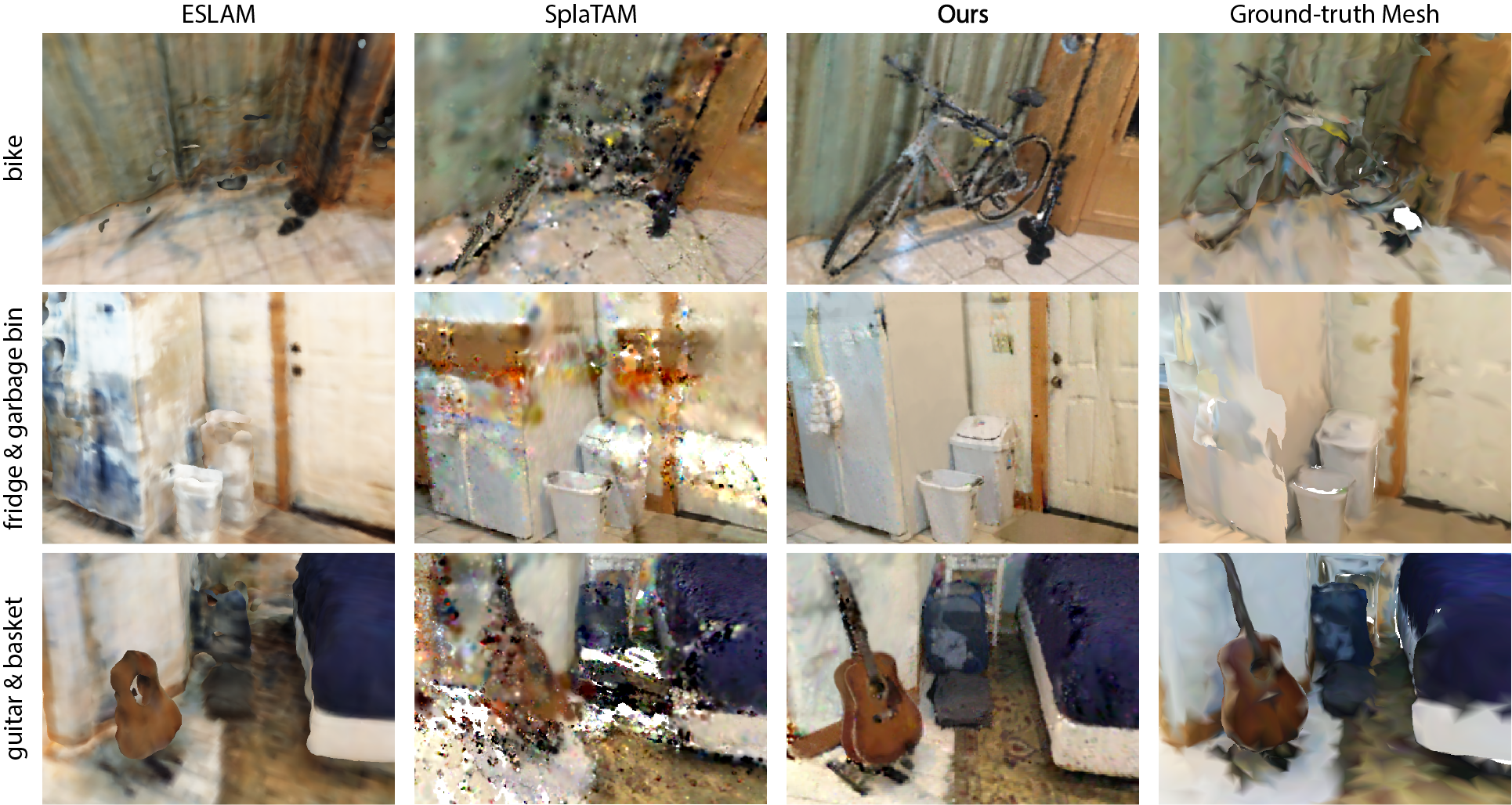}
    \end{center}
    \caption{The selected novel view synthesis of scene0000 from the ScanNet dataset \cite{dai2017scannet}. The rendered views display the reconstructed objects such as bike, fridge, garbage bin, and guitar from novel views. Our method outperforms baselines by a large margin primarily due to the integration of keyframe optimization and semantic constraints. Note that the ground-truth for novel views is captured from the offline-reconstructed mesh provided by the ScanNet dataset.}
    \label{fig:key_frame}
\end{figure}

In real-world datasets, tracking errors tend to accumulate along a trajectory, making pose estimations at later timestamps less reliable. Such inaccuracies can compromise the quality of map reconstructions, negatively impacting the previously well-established scenes. A case in point is scene0000 from the ScanNet dataset \cite{dai2017scannet}, where objects such as bike and guitar are revisited at early and late stages in the trajectory. Keyframes from later sequences, influenced by inaccurate camera poses, can disrupt the previously accurate reconstructions. \cref{fig:key_frame} illustrates the novel view evaluation for scene0000. In comparison to ESLAM \cite{johari2023eslam} and SplaTAM \cite{keetha2023splatam}, which are based on NeRF and 3D Gaussians, our method delivers more accurate reconstruction outcomes. The bike, garbage bin, and guitar are accurately rendered, meanwhile details are preserved. Our method facilitates the selection of keyframes based on geometric and semantic constraints, incorporating uncertainty weighting during the optimization of selected keyframes. This strategy demonstrates its effectiveness in map optimization from different views meanwhile preventing the unreliable keyframse with high uncertainty to significantly altering the earlier accurately reconstructed map.

\subsection{Scene Manipulation}
The obtained semantic mask within the 3D scene has a range of applications for subsequent tasks. As an illustrative example, we demonstrate a straightforward but efficient Gaussian editing method defined by \cref{eq:scene_manipulation}, which is crucial for enabling scene manipulation for robotics or mixed reality applications. 

\begin{figure}[tb]
    \begin{center}
        \includegraphics[width=0.8\textwidth]{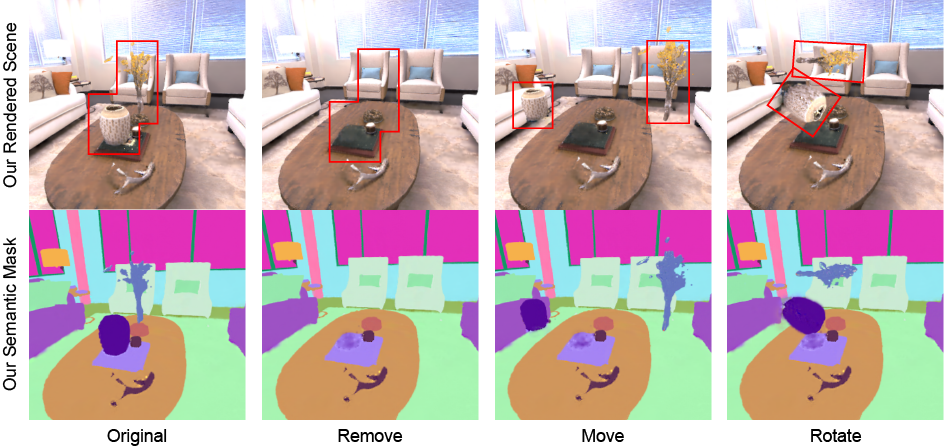}
    \end{center}
    \caption{The case study on scene manipulation in room0 of the Replica dataset \cite{straub2019replica}. We show the capabilities for object removal and transformation by specifying semantic labels. SGS-SLAM allows manipulation of either individual objects or a group of items, as illustrated by actions that include the removal of a jar and flowers, as well as moving and rotating them.}
    \label{fig:scene_edit}
\end{figure}

Utilizing the decoupled scene representation, in contrast to NeRF-based approaches that demand fine-tuning of the entire network, SGS-SLAM can edit specific objects within the scene while keeping the remainder of the well-trained, irrelevant environment fixed. As shown in \cref{fig:scene_edit}, we can directly manipulate the Gaussians associated with the editing target, such as erasing, moving, and rotating the jar and flowers on the table. In addition, we can group objects by selecting their semantic masks and applying a transition, such as rotating both the table and the above objects as shown in the supplementary material. This editing capability requires no training or fine-tuning, making it readily available for downstream applications.

\subsection{Ablation Study}
We perform the ablation of SGS-SLAM on the scene0000\_00 of the ScanNet dataset \cite{dai2017scannet} to evaluate the effectiveness of multi-channel feature supervision, and the keyframe optimization strategies.

\begin{table}[!h]
    \centering
    \caption{Ablation study of multi-channel optimization on the scene0000\_00 of the ScanNet dataset \cite{dai2017scannet}. The comparison involves settings where appearance, depth, and semantic supervision are removed. \xmark~means the metric is inapplicable.}
    \scriptsize 
    \begin{tabular*}{\textwidth}{@{\extracolsep{\fill}} l *{4}{c}}
        \Xhline{2\arrayrulewidth}
        \textbf{Settings} & \makecell{Depth L1 \\ {[cm]$\downarrow$}} & \makecell{ATE RMSE \\ {[cm]$\downarrow$}} & \makecell{PSNR \\ {[dB]$\uparrow$}} & \makecell{mIoU \\ {[\%]$\uparrow$}} \\ 
        \hline
        without color image ($C_{\rm pix}$) & 7.44 & 24.59 & \xmark & 68.19 \\
        without depth map ($D_{\rm pix}$) & 47.66 & 40.47 & 15.14 & 54.52 \\
        without semantic map ($S_{\rm pix}$) & 9.15 & 13.81 & 17.52 & \xmark \\
        without silhouette threshold ($Sil_{\rm pix}$) & 29.12 & 357.48 & 12.06 & 28.07 \\
        \hline
        with multi-channel optimization & \textbf{6.18} & \textbf{11.26} & \textbf{19.47} & \textbf{70.27} \\
        \Xhline{2\arrayrulewidth}
    \end{tabular*}
    \label{tab:ablation_channel}
\end{table}

\subsubsection{Effect of Multi-channel Optimization}

\begin{table}[h]
    \centering
    \caption{Ablation study of keyframe optimization on the scene0000 of the ScanNet dataset \cite{dai2017scannet}. The comparison involves settings where geometric, semantic, and uncertainty constraints are removed.}
    \scriptsize 
    \begin{tabular*}{\textwidth}{@{\extracolsep{\fill}} l *{4}{c}}
        \Xhline{2\arrayrulewidth}
        \textbf{Settings} & \makecell{Depth L1 \\ {[cm]$\downarrow$}} & \makecell{ATE RMSE \\ {[cm]$\downarrow$}} & \makecell{PSNR \\ {[dB]$\uparrow$}} & \makecell{mIoU \\ {[\%]$\uparrow$}} \\ 
        \hline
        without geometric threshold ($T_{\rm geo}$)& 6.66 & 15.55 & 19.21 & 68.93 \\
        without semantic threshold ($T_{\rm sem}$)& 8.44 & 12.89 & 17.84 & 69.85 \\
        without uncertainty weighting ($\mathcal{U}$)& 6.87 & 11.43 & 18.72 & 70.12 \\
        \hline
        with keyframe selection & \textbf{6.18} & \textbf{11.26} & \textbf{19.47} & \textbf{70.27} \\
        \Xhline{2\arrayrulewidth}
    \end{tabular*}
    \label{tab:ablation_keyframe}
\end{table}

\cref{tab:ablation_channel} shows the ablation study on multi-channel parameter optimization. The results reveal that our optimization strategy can significantly improve the localization and mapping performance. Specifically, the system without appearance color cannot provide rendered views, 
whereas camera pose and depth can still be estimated by leveraging depth and semantic input. The absence of depth data leads to the poorest depth estimation, highlighting the importance of geometric supervision. Furthermore, the absence of an input semantic map disables 3D semantic segmentation and remarkably diminishes the performance of tracking and mapping. Additionally, the silhouette threshold, essential for assessing scene visibility, is crucial for the system stability. Without this threshold, the system shows a significant decline in the effectiveness of tracking and mapping.

\subsubsection{Effect of Keyframe Optimization}

\cref{tab:ablation_keyframe} presents the results of keyframe selection ablation. Our two-level keyframe selection strategy reveals that omitting either geometric or semantic constraints results in a significant drop in both tracking and mapping performance. Additionally, without incorporating uncertainty weighting, the system demonstrates a decrease in performance compared to its full implementation. 

\section{Conclusion and Limitations}
We presented SGS-SLAM, the first semantic dense visual SLAM system based on the 3D Gaussian representation. We propose to leverage multi-channel parameter optimization where appearance, geometric, and semantic constraints are combined to enforce high-accurate 3D semantic segmentation, and high-fidelity dense map reconstruction meanwhile effectively producing a robust camera pose estimation. SGS-SLAM takes advantage of optimal keyframe optimization, resulting in reliable reconstruction quality. Extensive experiments show that our method provides state-of-the-art tracking and mapping results, meanwhile maintaining rapid rendering speeds. Furthermore, the high-quality reconstruction of scenes and precise 3D semantic labeling generated by our system establish a strong foundation for downstream tasks such as scene editing, offering solid prior for robotics or mixed reality applications. 

\subsubsection{Limitations}
SGS-SLAM replies on depth and 2D semantic signal inputs for tracking and mapping. In scenarios where this information is scarce or difficult to access, the system's effectiveness will be compromised. Additionally, our method incurs large memory consumption when deployed to large scenes. Addressing these limitations will be an objective for future research.


%
%
\bibliographystyle{splncs04}
\bibliography{egbib}

\newpage
{
    \centering
    \bfseries\Large SGS-SLAM: Semantic Gaussian Splatting For Neural Dense SLAM\\[10pt]
    \bfseries\large — Supplementary Material —
    \par
}
\vspace{2em}

\section{Experiment Settings}

In this section, we outline the experimental setup and hyperparameters applied in our studies. The experiments were conducted on a server with NVIDIA A100-40GB GPU. However, our method typically takes less than 12 GB of memory for the scenes presented in this study, making it compatible with any GPU that has more than this amount of memory. The ground-truth results we compared, particularly for the novel view rendering, were obtained from the ground-truth mesh provided in the dataset, which was generated in an offline manner. Therefore, some defects can be observed in the ground-truth results. The code will be released soon.

\subsubsection{SGS-SLAM}
By default, both mapping and tracking operations are conducted for each frame. During the tracking phase, we set the silhouette visibility threshold, $T_{\rm sil}$, to 0.99. The multi-channel optimization involves three parameters: $\lambda_D = 1.0$ for depth, $\lambda_C = 0.5$  for colors, and $\lambda_S = 0.05$ for semantic loss, with the semantic loss weight being comparatively low due to the typical noisiness of real-world semantic labels. Throughout the tracking, the multi-channel Gaussian parameters remain constant, adjusting only the camera parameters with a learning rate of 2e-3 for transition. Key-frames are initially chosen at intervals of every 5 frames, then refined based on geometric and semantic criteria. The geometric overlap threshold, $\eta$, is defined at 0.05, and the semantic mean Intersection over Union (mIoU) threshold, $T_{\rm sem}$, at 0.7. The maximum number of keyframes per frame is limited to 25, considering the computation speed. The uncertainty decay coefficient, $\tau$ scales with the length of the input frame series. In the mapping process, the silhouette threshold $T_{\rm sil}$ is adjusted to 0.5. The weights of photometric loss are set to $\lambda_D = 1.0$, $\lambda_C = 0.5$, and $\lambda_S = 0.1$. Here, camera parameters are fixed, and Gaussian parameters are optimized, with specific learning rates for 3D position at 1e-4, color 2.5e-3, Gaussian rotation at 1e-3, logit opacity at 0.05, and log scale at 1e-3. Performance metrics of tracking and mapping are assessed every 5 frames, with mIoU scores evaluated at the same frequency. 

The mapping and tracking iteration steps are specific to each dataset, In the case of the Replica dataset \cite{straub2019replica}, the number of iterations for tracking and mapping are set to 40 and 60. For the ScanNet dataset \cite{dai2017scannet}, tracking and mapping are set to 120 and 40. In the enhanced ScanNet++ dataset \cite{yeshwanthliu2023scannetpp}, where the camera transition is large between each frame, the tracking and mapping iterations are adjusted to 220 and 50.

\subsubsection{Baselines}
We adhere to the default configurations for each baseline as reported in their papers. The evaluation metrics for tracking and mapping are consistent with those applied to our method. For baselines whose implementations are not publicly available, we present the results as reported in their papers.

\begin{table}[th]
    \centering
    \caption{Quantitative comparison of ATE RMSE [cm] between our method and the baselines for each scene of the Replica dataset \cite{straub2019replica}. Our method demonstrates \colorbox{lightblue}{SOTA} performances.}
    \scriptsize 
    \begin{tabularx}{\textwidth}{@{}l@{\hspace{4pt}} | *{9}{>{\centering\arraybackslash}X}@{}}
        \Xhline{2\arrayrulewidth}
        \textbf{Methods} & Avg. & Room0 & Room1 & Room2 & Office0 & Office1 & Office2 & Office3 & Office4 \\ 
        \hline
        Vox-Fusion & 3.09 & 1.37 & 4.70	& 1.47 & 8.48 & 2.04 & 2.58 & 1.11 & 2.94 \\
        NICE-SLAM & 2.50 & 2.25 & 2.86 & 2.34 & 1.98 & 2.12 & 2.83 & 2.68 & 2.96 \\
        Co-SLAM & 0.86 & 0.65 & 1.13 & 1.43 & 0.55 & 0.50 & 0.46 & 1.40 & 0.77 \\
        ESLAM & 0.63 & 0.71 & 0.70 & 0.52 & 0.57 & 0.55 & 0.58 & 0.72 & 0.63 \\
        Point-SLAM & 0.52 & 0.61 & 0.41 & 0.37 & 0.38 & 0.48 & 0.54 & 0.69 & 0.72 \\
        \textbf{Ours} & \cellcolor{lightblue}\textbf{0.41} &
        \cellcolor{lightblue}\textbf{0.46} &
        \cellcolor{lightblue}\textbf{0.45} &
        \cellcolor{lightblue}\textbf{0.29} &
        \cellcolor{lightblue}\textbf{0.46} &
        \cellcolor{lightblue}\textbf{0.23} &
        \cellcolor{lightblue}\textbf{0.45} &
        \cellcolor{lightblue}\textbf{0.42} &
        \cellcolor{lightblue}\textbf{0.55} \\
        \Xhline{2\arrayrulewidth}
    \end{tabularx}
    \label{tab:sup_ate_rmse_replica}
\end{table}

\begin{table}[th]
    \centering
    \caption{Quantitative comparison of ATE RMSE [cm] between our method and the baselines for the selected scenes on the ScanNet dataset \cite{dai2017scannet}.}
    \scriptsize 
    \begin{tabularx}{\textwidth}{@{}l@{\hspace{4pt}} | *{7}{>{\centering\arraybackslash}X}@{}}
        \Xhline{2\arrayrulewidth}
        \textbf{Methods} & Avg. & 0000 & 0059 & 0106 & 0169 & 0181 & 0207 \\ 
        \hline
        Vox-Fusion & 26.90 & 68.84 & 24.18 & 8.41 & 27.28 & 23.30 & 9.41 \\
        NICE-SLAM & 10.70 & 12.00 & 14.00 & 7.90 & 10.90 & 13.40 & 6.20 \\
        Co-SLAM & 9.73 & 12.29 & 9.57 & \cellcolor{lightblue}\textbf{6.62} & 13.43 & \cellcolor{lightblue}\textbf{7.13} & 9.37 \\
        ESLAM & \cellcolor{lightblue}\textbf{7.88} & \cellcolor{lightblue}\textbf{8.47} & 8.70 & 7.58 & \cellcolor{lightblue}\textbf{7.45} & 8.87 & 6.20 \\
        Point-SLAM & 12.19 & 10.24 & \cellcolor{lightblue}\textbf{7.81} & 8.65 & 22.16 & 14.77 & 9.54 \\
        \textbf{Ours} & 9.87 & 11.15 & 9.54 & 10.43 & 10.70 & 11.28 & \cellcolor{lightblue}\textbf{6.11} \\
        \Xhline{2\arrayrulewidth}
    \end{tabularx}
    \label{tab:sup_ate_rmse_scannet}
\end{table}

\begin{table}[th]
    \centering
    \caption{Quantitative comparison of ATE RMSE [cm] between our method and the baseline for the selected scenes on the ScanNet++ dataset \cite{yeshwanthliu2023scannetpp}.}
    \scriptsize 
    \begin{tabularx}{\textwidth}{@{}l@{\hspace{15pt}} | *{3}{>{\centering\arraybackslash}X}@{}}
        \Xhline{2\arrayrulewidth}
        \textbf{Methods} & Avg. [cm]$\downarrow$ & 8b5caf3398 [cm]$\downarrow$ & b20a261fdf [cm]$\downarrow$ \\ 
        \hline
        ESLAM & 170.06 & 185.15 & 156.96 \\
        \textbf{Ours} & \cellcolor{lightblue}\textbf{1.62} & \cellcolor{lightblue}\textbf{0.65} & \cellcolor{lightblue}\textbf{2.34} \\
        \Xhline{2\arrayrulewidth}
    \end{tabularx}
    \label{tab:sup_ate_rmse_scannetpp}
\end{table}

\section{Additional Experiment Results}
We provide additional quantitative analysis of camera tracking in \cref{sec:camera_tracking}. The visualization of semantic segmentation compared with NeRF-based method is presented in \cref{sec:training_seg}. More qualitative novel view rendering results are illustrated in \cref{sec:novel_view}. We compared our method with Vox-Fusion \cite{yang2022vox}, NICE-SLAM \cite{zhu2022nice}, Co-SLAM \cite{wang2023co}, ESLAM \cite{johari2023eslam}, and Point-SLAM \cite{sandstrom2023point} for ATE RMSE evaluation. For 3D semantic segmentation, we visualized the comparison with DNS-SLAM \cite{li2023dns}.

\subsection{Camera Tracking}
\label{sec:camera_tracking}

In this section, we break down the quantitative analysis on ATE RMSE [cm] on Replica \cite{straub2019replica}, ScanNet \cite{dai2017scannet}, and ScanNet++ \cite{yeshwanthliu2023scannetpp} datasets. \cref{tab:sup_ate_rmse_replica}, \cref{tab:sup_ate_rmse_scannet}, and \cref{tab:sup_ate_rmse_scannetpp} present the evaluation our SGS-SLAM against baseline models on each dataset. Our method of estimating camera poses by directly optimizing the gradient on dense photometric loss achieves state-of-the-art tracking performance on datasets with high-quality RGB-D images. In particular, on the ScanNet++ dataset \cite{yeshwanthliu2023scannetpp}, where there is a large camera transition between successive frames, NeRF-based methods like ESLAM failed to track. Conversely, SGS-SLAM demonstrated robust and accurate tracking capability.

\subsection{Semantic Segmentation}
\label{sec:training_seg}

\begin{figure}[t]
    \begin{center}
        \includegraphics[width=0.9\textwidth]{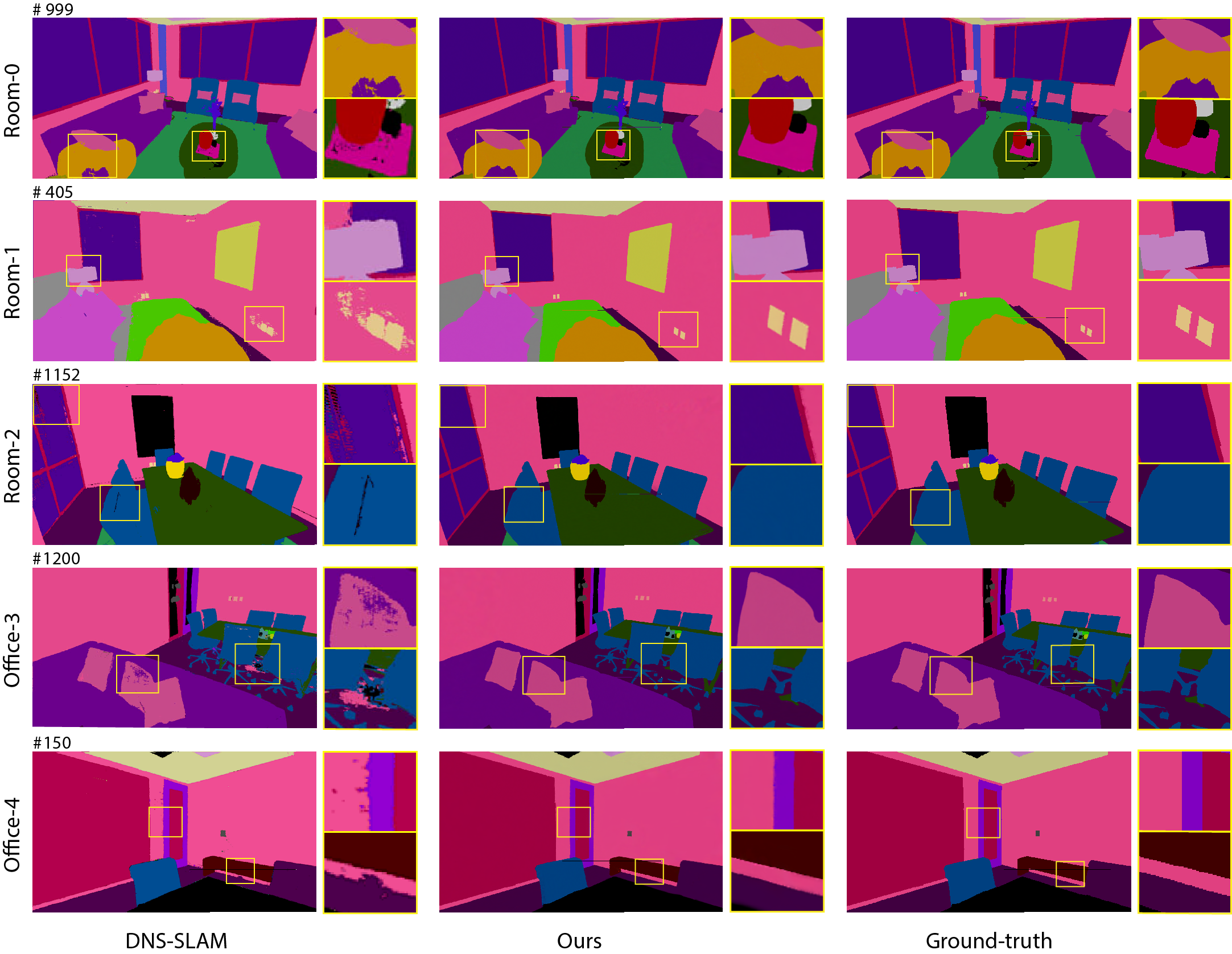}
    \end{center}
    \caption{Qualitative comparison of our method and DNS-SLAM \cite{li2023dns} for semantic segmentation from the Replica dataset \cite{straub2019replica}. The visualization outcomes of DNS-SLAM \cite{li2023dns} are obtained from its paper. The frames of the training view are chosen based on the results presented in DNS-SLAM. Compared to NeRF-based models, our approach delivers segmentation results with higher accuracy.}
    \label{fig:sup_replica_semantic}
\end{figure}

In this section, the outcomes of semantic segmentation on the Replica dataset \cite{straub2019replica} are visualized and compared with DNS-SLAM \cite{li2023dns}, a NeRF-based approach. As illustrated in \cref{fig:sup_replica_semantic}, our method offered accurate and detailed segmentation, whereas DNS-SLAM faces challenges in edges due to the over-smoothing issue of NeRF.

\subsection{Novel View Rendering}
\label{sec:novel_view}

\begin{figure}[H]
    \begin{center}
        \includegraphics[width=0.83\textwidth]{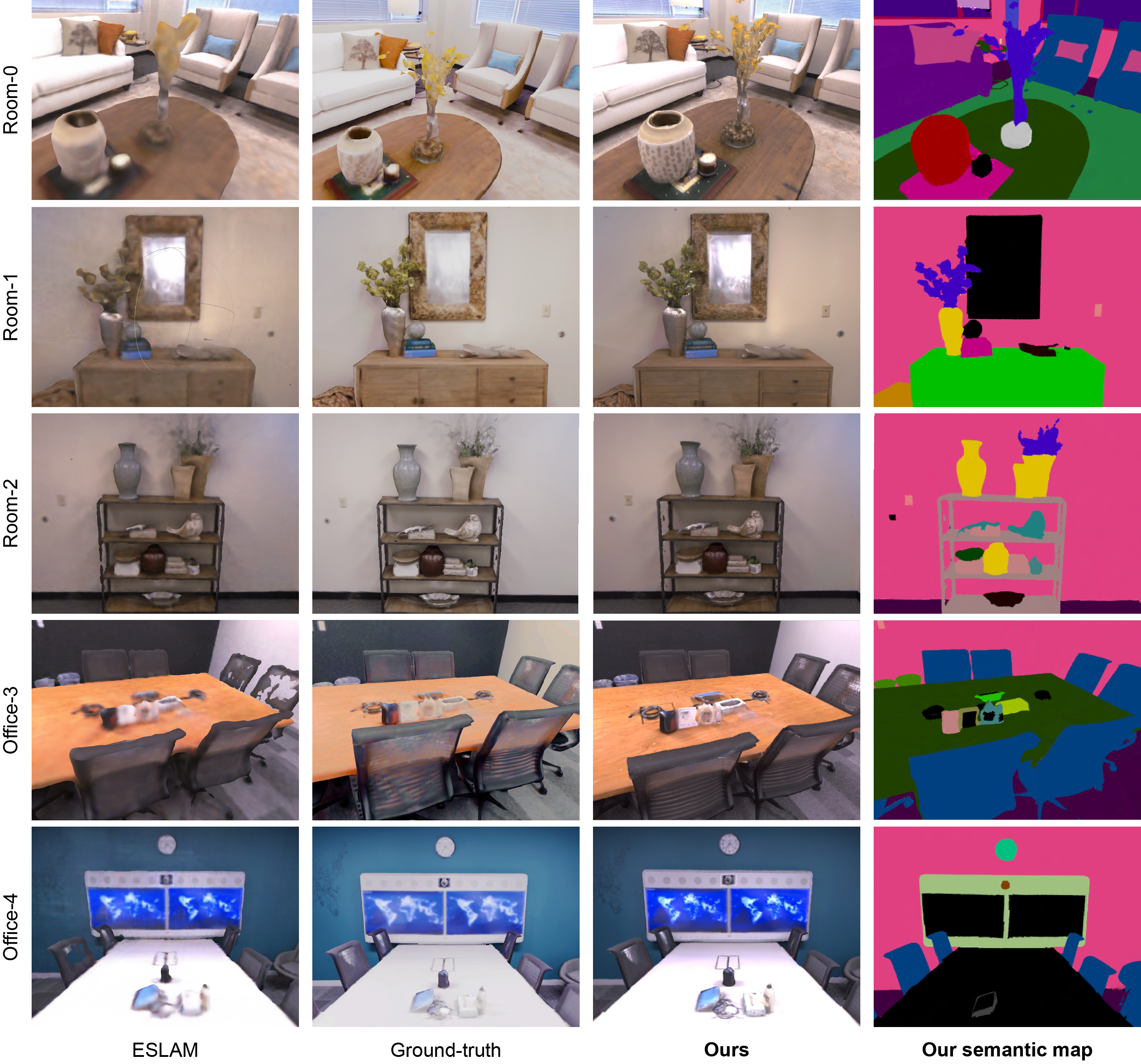}
    \end{center}
    \caption{The visualization of novel view rendering between the ESLAM \cite{johari2023eslam} and our method on the Replica dataset \cite{straub2019replica}. The ground-truth novel views are captured from meshes provided by the dataset.}
    \label{fig:sup_replica}
\end{figure}

\begin{figure}[H]
    \begin{center}
        \includegraphics[width=0.83\textwidth]{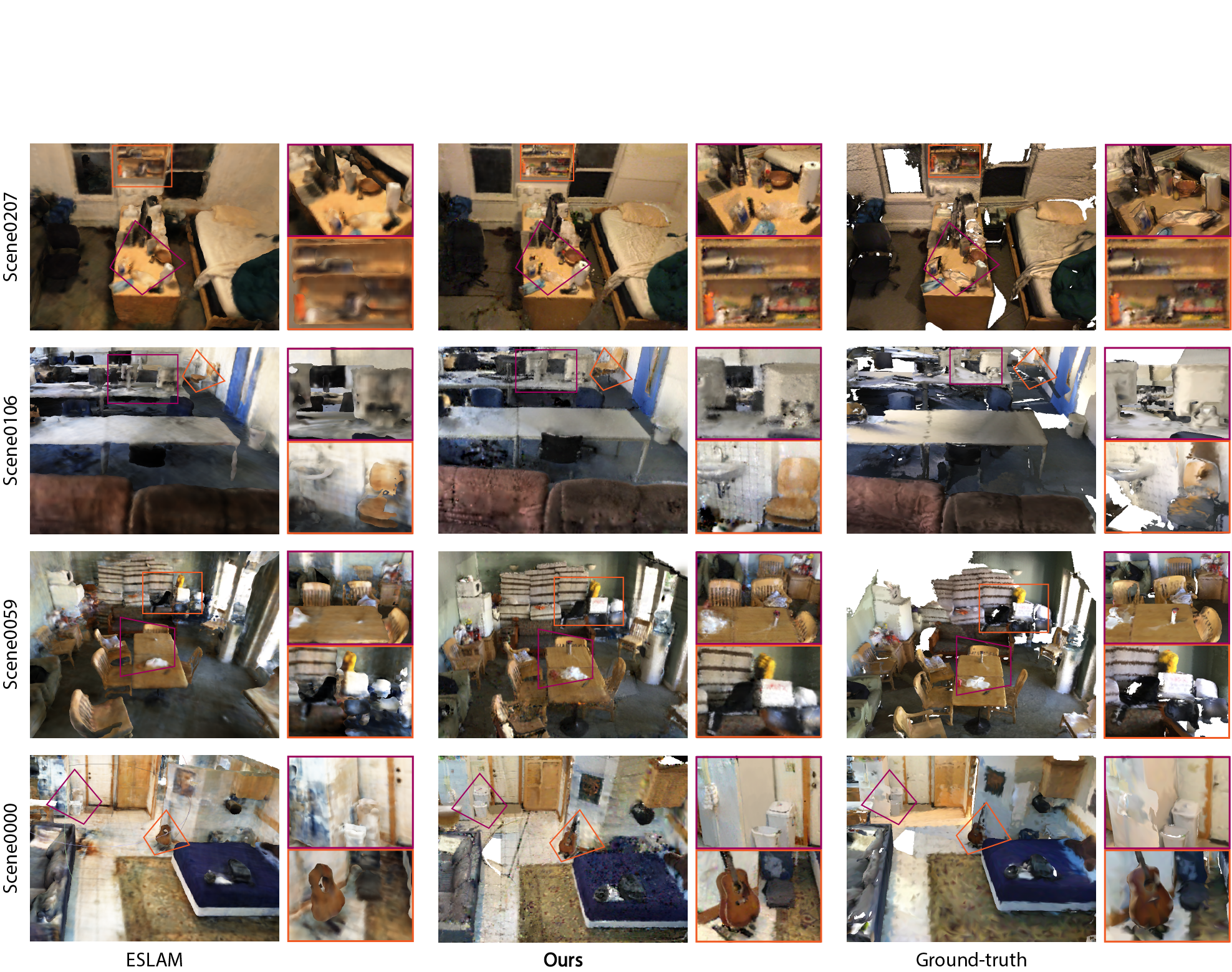}
    \end{center}
    \caption{The visualization of novel view rendering between the baseline and our method using the ScanNet dataset \cite{dai2017scannet}. The ground-truth novel views are captured from meshes. SGS-SLAM exhibits rendering of high fidelity and outperforms the NeRF-based ESLAM \cite{johari2023eslam}. In contrast to the ground-truth mesh, our method demonstrates robust mapping in areas where the ground-truth mesh presents holes.}
    \label{fig:sup_scannet}
\end{figure}

\begin{figure}[H]
    \begin{center}
        \includegraphics[width=0.83\textwidth]{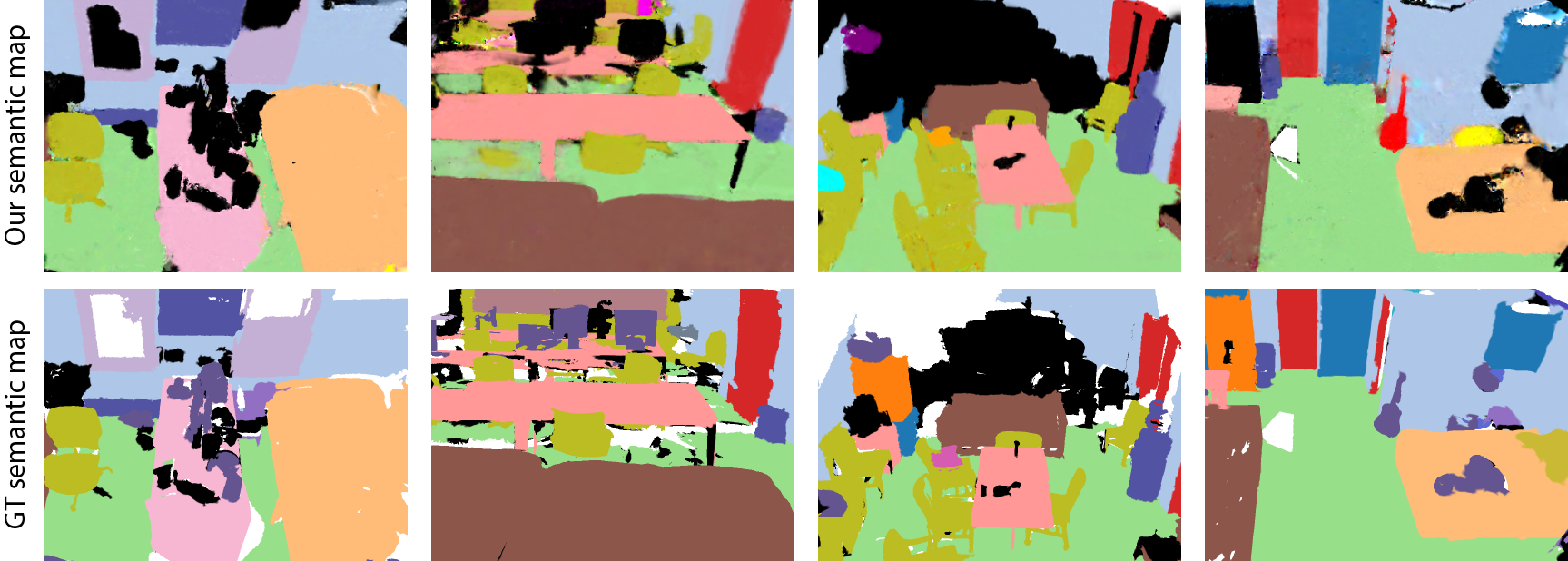}
    \end{center}
    \caption{The visualization of 3D semantic segmentation results of SGS-SLAM, as applied to the novel views selected in \cref{fig:sup_scannet}. Note that the rendering results exhibit minor variations in scene objects due to the use of a modified semantic dataset from ScantNet. For our method, the training data is processed from the filtered semantic labels using the nyu40-class, where certain objects are not distinctly labeled and are assigned as background (depicted in black). Furthermore, we introduce extra labels, like guitar, bag, and basket, to enhance the quality of scene reconstruction.}
    \label{fig:sup_scannet_semantic}     
\end{figure}

\begin{figure}[H]  
    \begin{center}
        \includegraphics[width=0.9\textwidth]{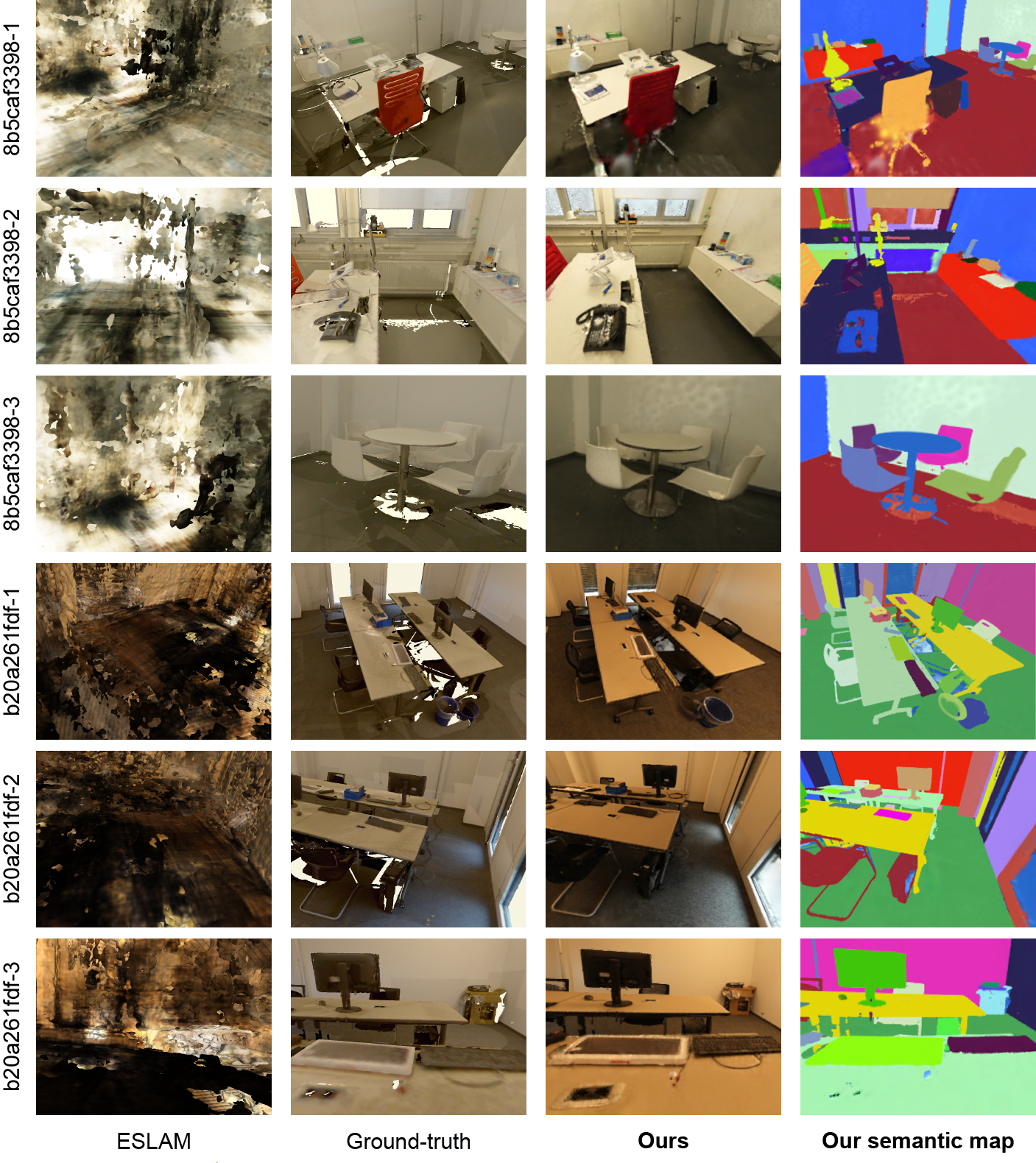}
    \end{center}
    \caption{The visualization of novel view rendering between the baseline and our method using the ScanNet++ dataset \cite{yeshwanthliu2023scannetpp}. The ground-truth novel views are captured from meshes. SGS-SLAM demonstrates superior rendering quality, while ESLAM \cite{johari2023eslam} suffers from significant tracking errors and fails to reconstruct the map. In addition, our method also offers accurate instance-level segmentation outcomes.}
    \label{fig:sup_scannetpp}
\end{figure}

We present additional results of novel view rendering using our method across the Replica \cite{straub2019replica}, ScanNet \cite{dai2017scannet}, and ScanNet++ \cite{yeshwanthliu2023scannetpp} datasets, with comparisons to ESLAM \cite{johari2023eslam}. Visualizations are provided in \cref{fig:sup_replica}, \cref{fig:sup_scannet}, \cref{fig:sup_scannet_semantic}, and \cref{fig:sup_scannetpp} with semantic segmentation outcomes. Our method consistently delivers high-quality rendering results for both synthesized and real-world datasets. Notably, on the challenging real-world ScanNet++ dataset, ESLAM \cite{johari2023eslam} struggled to reconstruct the scene. By contrast, SGS-SLAM provides accurate high-fidelity scene reconstructions along with precise segmentation outcomes. Note that the ground-truth segmentation labels are retrieved from the ground-truth mesh at the instance level, and therefore, our results also show instance-level segmentation.

\subsection{Scene Manipulation}

\begin{figure}[h]  
    \begin{center}
        \includegraphics[width=0.9\textwidth]{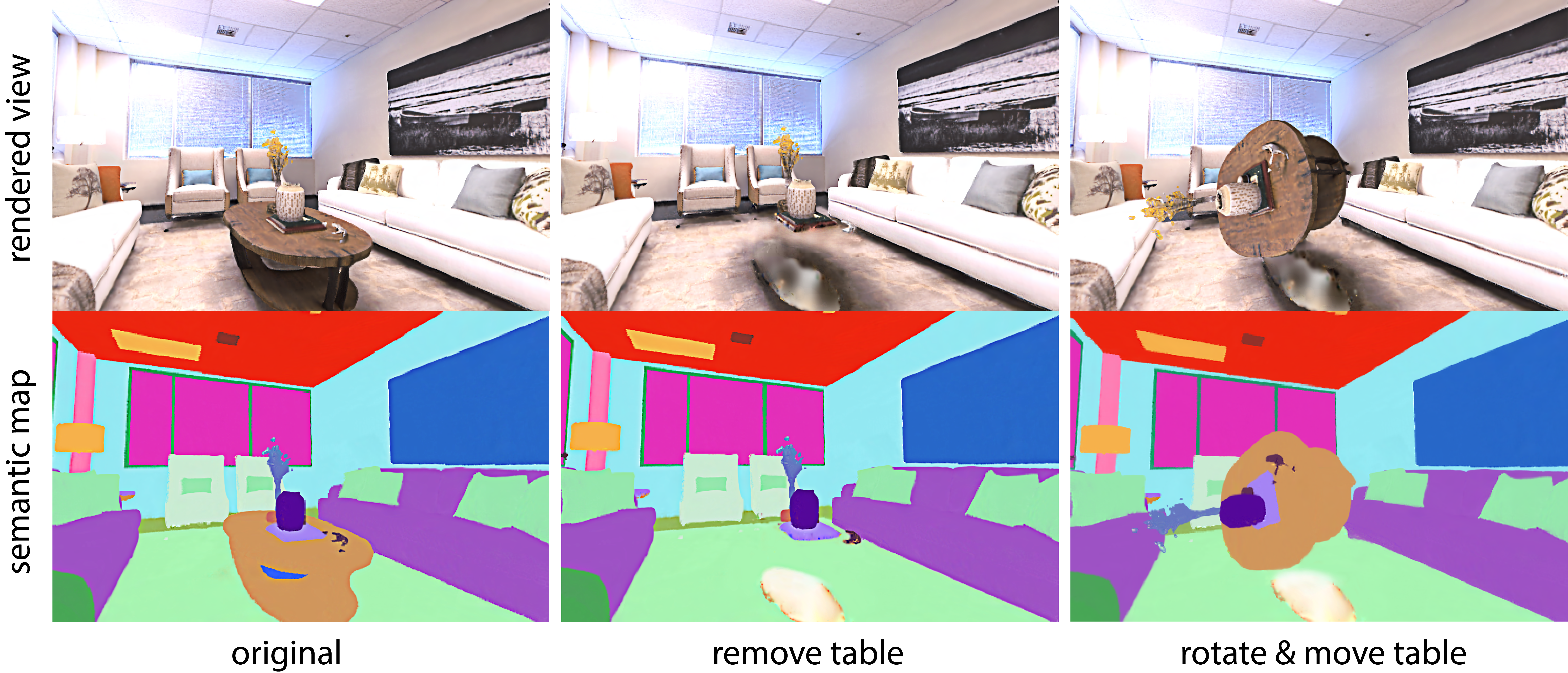}
    \end{center}
    \caption{The visualization of scene manipulation by grouping Gaussians via semantic labels. SGS-SLAM allows manipulation of either individual objects or a group of items, as illustrated by actions that include the removal of a table, as well as moving and rotating the table together with all objects on it.}
    \label{fig:sup_scene_edit}
\end{figure}

In this section, we visualize scene manipulation results by grouping the Gaussians using the semantic mask. As shown in \cref{fig:sup_scene_edit}, for object removal, we can directly erase the Gaussians associated with the editing target, such as removing the table while preserving all the items on it. In addition, we can group objects by selecting their semantic masks and applying translation and rotation, such as moving and rotating both the table and the above objects to a different place.

It is worth noting that we can observe holes left in the place when removing or transitioning the objects. Such as the hole left on the ground when we removed the table. This is due to the explicit scene representation using 3D Gaussians where the unobserved geometry in the multi-views from the trajectory are inevitably missing. This defect, stemming from the characteristics of the 3D Gaussian representation, poses a challenging problem. It is identified as an area for future research, with the potential solution through the use of 3D geometry priors \cite{freda2023plvs} or scene inpainting \cite{ye2023gaussian} techniques.

\end{document}